\documentclass[letterpaper, 10 pt, conference]{ieeeconf}  

\IEEEoverridecommandlockouts                              

\usepackage{graphics} 
\usepackage{epsfig} 
\usepackage{mathptmx} 
\usepackage{times} 
\usepackage{amsmath} 
\usepackage{amssymb}  
\usepackage{kotex}
\usepackage{indentfirst}
\usepackage{subfigure}
\usepackage{algorithm,algpseudocode}
\usepackage{accents}

\usepackage[table,xcdraw]{xcolor}

\DeclareMathOperator*{\argmin}{arg\,min}

\newcommand\uubar[1]{\underaccent{\bar}{#1}}
\newtheorem{definition}{Definition}
\newtheorem{assumption}{Assumption}

\title{\LARGE \bf
Reachable Set-based Path Planning for
Automated Vertical Parking System}

\author{In~Hyuk~Oh$^{1}$, Ju~Won~Seo$^{1}$, Jin~Sung~Kim$^{1}$, and Chung~Choo~Chung$^{2}$$^\dag$
\thanks{*This work was supported by the National
Research Foundation of Korea (NRF) through the Korea Government Ministry
of Science and Information and Communication Technology (MSIT), Data Driven Optimized Autonomous Driving Technology Using Open Set Classification Method, under Grant 2021R1A2C2009908.}
\thanks{In Hyuk Oh, Ju Won Seo, and Jin Sung Kim are with Dept. of Electrical Engineering, Hanyang University, Seoul 04763, Korea (email : \{dhdlsgur12, suhju1227, jskim06\}@hanyang.ac.kr)}%
\thanks{C. C. Chung is with Div. of Electrical and Biomedical Engineering, Hanyang University, Seoul 04763, Korea (email: cchung@hanyang.ac.kr)}
\thanks{\dag: Corresponding author}%
}

\begin{document}

\maketitle
\thispagestyle{empty}
\pagestyle{empty}

\begin{abstract}
This paper proposes a local path planning  method with a reachable set for Automated vertical Parking Systems (APS). First, given a parking lot layout with a goal position, we define an intermediate pose for the APS to accomplish reverse parking with a single maneuver, i.e.,  without changing the gear shift. Then, we introduce a reachable set which is a set of points consisting of the grid points of all possible intermediate poses. Once the APS approaches the goal position, it must select an intermediate pose in the reachable set. A minimization problem was formulated and solved to choose the intermediate pose.
We performed various scenarios with different parking lot conditions. We used the Hybrid-A* algorithm for the global path planning to move the vehicle from the starting pose to the intermediate pose and utilized clothoid-based local path planning to move from the intermediate pose to the goal pose. Additionally, we designed a controller to follow the generated path and validated its tracking performance. It was confirmed that the tracking error in the mean root square for the lateral position was bounded within   0.06m and for orientation within 0.01rad.
\end{abstract}

\section{INTRODUCTION}
An automated parking system (APS) has recently received significant attention in the autonomous system field. Automated parking system typically consists of environment perception of the parking lot, vehicle localization, path planning, and path tracking. Path planning is an essential technology for implementing collision-free and safe parking in the APS. Path planning finds a trajectory the vehicle can follow from a starting position to a goal position while avoiding obstacles or driving in a narrow space. State-of-the-art approaches for path planning are reported in~\cite{c1}. Path planning can be categorized into global path planning and local path planning. Global path planning algorithms such as Hybrid-A*~\cite{c2}, rapidly-exploring random trees (RRT*)~\cite{c3}, and folding-based path planning~\cite{c4} solve the obstacle avoidance problem considering the entire configuration including the starting and goal position. On the other hand, local path planning can perform path corrections and improvements during robot motion to cope with obstacles position or environmental changes that the robot faces~\cite{c5}. Then, possible motion tasks, such as path following, are assigned for suitable feedback controllers in a car-like robot~\cite{c6}.

Generally, geometric paths are used for path following, considering the car-like robot's steering wheel angle constraints~\cite{c6}.
The geometric path uses lines, circles, polynomials, and clothoid curves for automated vertical parking. Using circle-to-circle and circle-line-circle algorithms for local path planning demonstrates exemplary performance in terms of computation time. However, the results path may not be smooth at certain moments, and the steering wheel angle required to follow it becomes discontinuous~\cite{c7}. There has been a study on generating parking paths using the clothoid curve, calculating using Fresnel integrals to prevent the discontinuity of steering wheel angle~\cite{c8,c9}. However, computing the local path planning through integration at every sample time requires significant computation time, so alternative approaches are needed. An approximated clothoid-based local path planning method was recently proposed to save unnecessary computation time~\cite{c10}. However, this method may not be able to generate a path depending on the starting pose for reverse parking, so selecting an appropriate pose is necessary.

The concept of a reachable set can be used to determine the pose of an ego vehicle for a vertical parking maneuver \cite{c10}. The reachable set is the set of all poses that can be reached to the goal pose using local path planning. To utilize the reachable set in an automated vertical parking system, the vehicle's pose is divided into three categories: starting, intermediate, and goal pose~\cite{c12,c13}. The starting pose is where the APS algorithm begins, the intermediate pose is where the vehicle begins the vertical parking maneuver, and the goal pose is the end of the vertical parking maneuver. A key aspect to consider here is that intermediate poses significantly impact the feasibility of reaching the goal pose. Therefore, determining the intermediate pose is essential for completing the vertical parking maneuver so that the reachable set can be used to select the appropriate intermediate pose.

This paper proposes a path planning method using the clothoid-based reachable set for an automated vertical parking system. We generate two paths: the path from the starting pose to the intermediate pose and the path from the intermediate pose to the goal pose.
The former path is generated using the Hybrid-A* algorithm since Hybrid-A* can accurately match the specified pose even in tight spaces~\cite{c13}. This global path planning considers obstacles and the surrounding environment of the parking lot.
The latter path, which requires accurate reaching of the parking spot, is generated using approximated clothoid-based path planning.
To find an appropriate intermediate pose, the reachable set concept is used. The reachable set is a 3-dimensional grid point with longitudinal and lateral positions and orientation of intermediate poses where clothoid-based path generation is possible up to the final parking spot.
The reachable set generation algorithm can conduct a collision-free reachable set considering obstacles. Then, a cost function is defined to select an appropriate intermediate pose. Finally, a system for path planning and tracking from the starting pose to the goal pose is designed. To show the effectiveness of the proposed method, numerical simulations were conducted for cases with and without obstacles, with parking corridor widths of 7m and 6m. From the simulation results, it is shown that a successful path was generated without collisions with obstacles. In addition, the root mean square error (RMSE) for tracking, the performance of the lateral and orientation was verified to be 0.02m and 0.006rad for the 7m case, 0.06m and 0.01rad for 6m case, respectively.

\section{Approach to Automated Parking System}

\begin{figure}[t]
\centering
\includegraphics[width=0.4\textwidth]{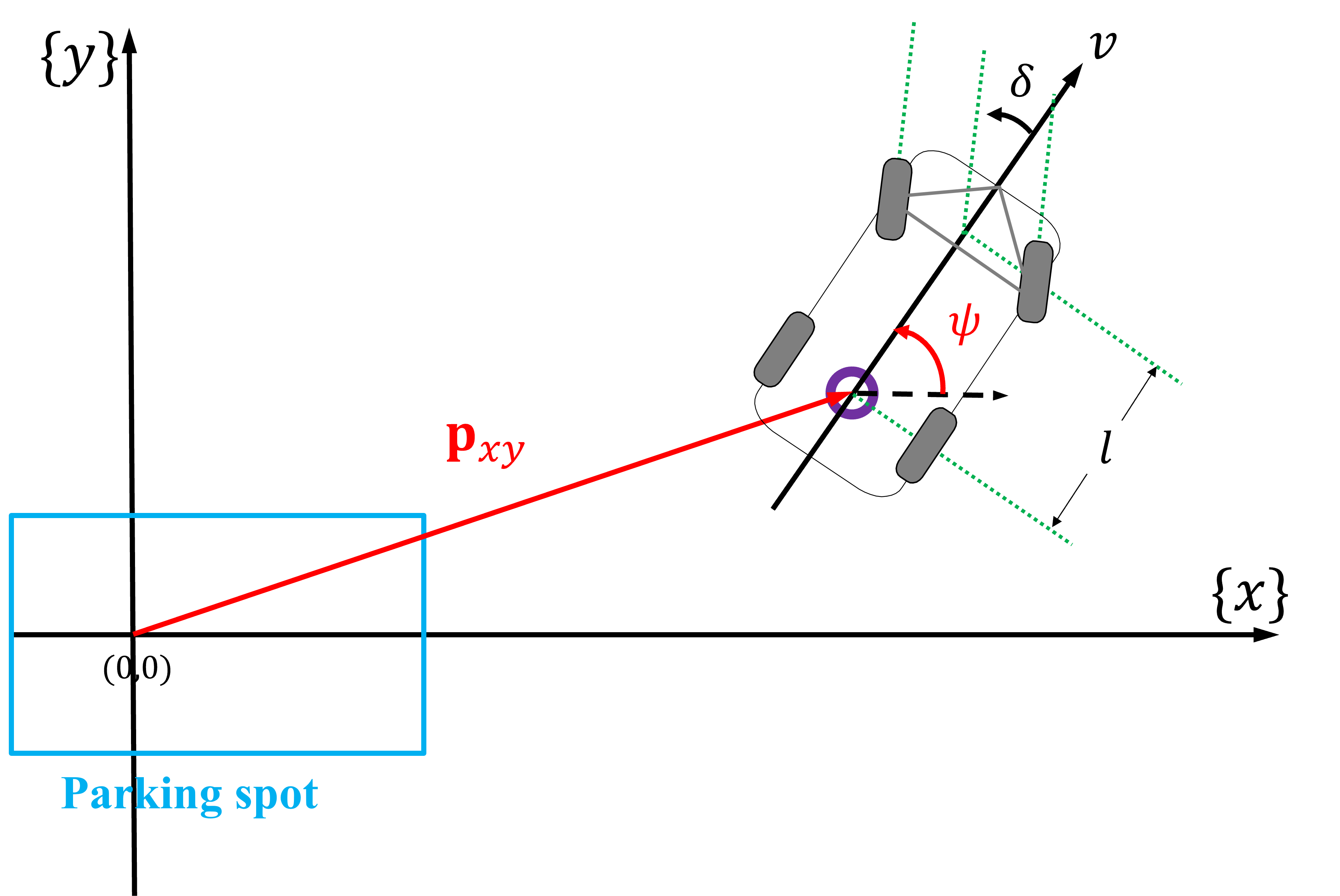}   
\caption{Parking coordinates and nonholonomic vehicle model at the center of the rear axle ($cra$) of the vehicle.}
\label{fig:coordinate}
\end{figure}

\subsection{Vehicle Kinematic Model}

Let us consider the parking coordinates $\{x\,y\}$ described in Fig.1. We define the vehicle position vector ${\mathbf p}_{xy}=[x,\, y]^T\in \mathbb{R}^2$ at the center of the rear axle $(cra)$ in the parking coordinates. Then, the vehicle pose at the $cra$ by ${\mathbf p}_{cra}\in\mathbb{R}^3$ is defined with the orientation of the vehicle $\psi$ by

\begin{gather}
\label{eq:state}
{\mathbf p}_{cra}= \begin{bmatrix}{\mathbf p}_{xy}\\\psi\end{bmatrix}=\begin{bmatrix}x\\y\\\psi\end{bmatrix}.
\end{gather}
This study assumes the Ackerman turning geometry and the bicycle model~\cite{c14,c15}. In addition, we assume that the tire slip angle of the vehicle is neglected~\cite{c15}. Then, for lateral vehicle motion, we use a discrete-time kinematic model and the process of converting from a continuous-time kinematic model, which is derived in~\cite{c14}. The vehicle kinematic model can be obtained using a zero-order hold with sampling rate $T_{s}$ as follows:

\begin{subequations}
\label{eq:discrete_kinematic_model}
\begin{equation}
\label{eq:lon_model}
x(k+1)=x(k) + T_{s} v(k) \cos{(\psi(k))}
\end{equation}
\begin{equation}
\label{eq:lat_model}
\begin{split}
y(k+1)&=y(k) + T_{s} v(k) \sin{(\psi(k))} \\
\psi(k+1)&=\psi(k) + T_{s}\frac{v(k)}{l} \tan{(\delta(k))}
\end{split}
\end{equation}
\end{subequations}
where $v(\cdot),\delta(\cdot),$and $l$ denote the longitudinal velocity at the $cra$, the front wheel steering angle, and the wheelbase, respectively.

\newtheorem{remark}{Remark}
\begin{remark}
Note that~\eqref{eq:discrete_kinematic_model} is separated into the longitudinal motion~\eqref{eq:lon_model} and the lateral motion model~\eqref{eq:lat_model} for the decentralized controller~\cite{c9,c10}. The desired longitudinal velocity is designed by $v^d(k)=-\nu x(k)$ for any $\nu \in (0,2/T_s)$~\cite{c9,c10}, and a basic feedback control law, such as PD control, is used to track this velocity. And we consider simplified model of the power train dynamics given by
$\ddot{x}(k+1)=(1-\frac{T_s}{\tau})\ddot{x}(k)+\frac{T_s}{\tau}\ddot{x}^d(k)$
with a time constant $\tau$~\cite{c15}. In addition, the magnitude of the desired longitudinal velocity $|v^d(\cdot)|$ is limited to account for the vehicle kinematic motion during parking operation. Readers who are interested in the detailed background and description of the longitudinal motion can be referred to \cite{c10}.
\end{remark}
\begin{figure}[t]
\centering
\includegraphics[width=0.4\textwidth]{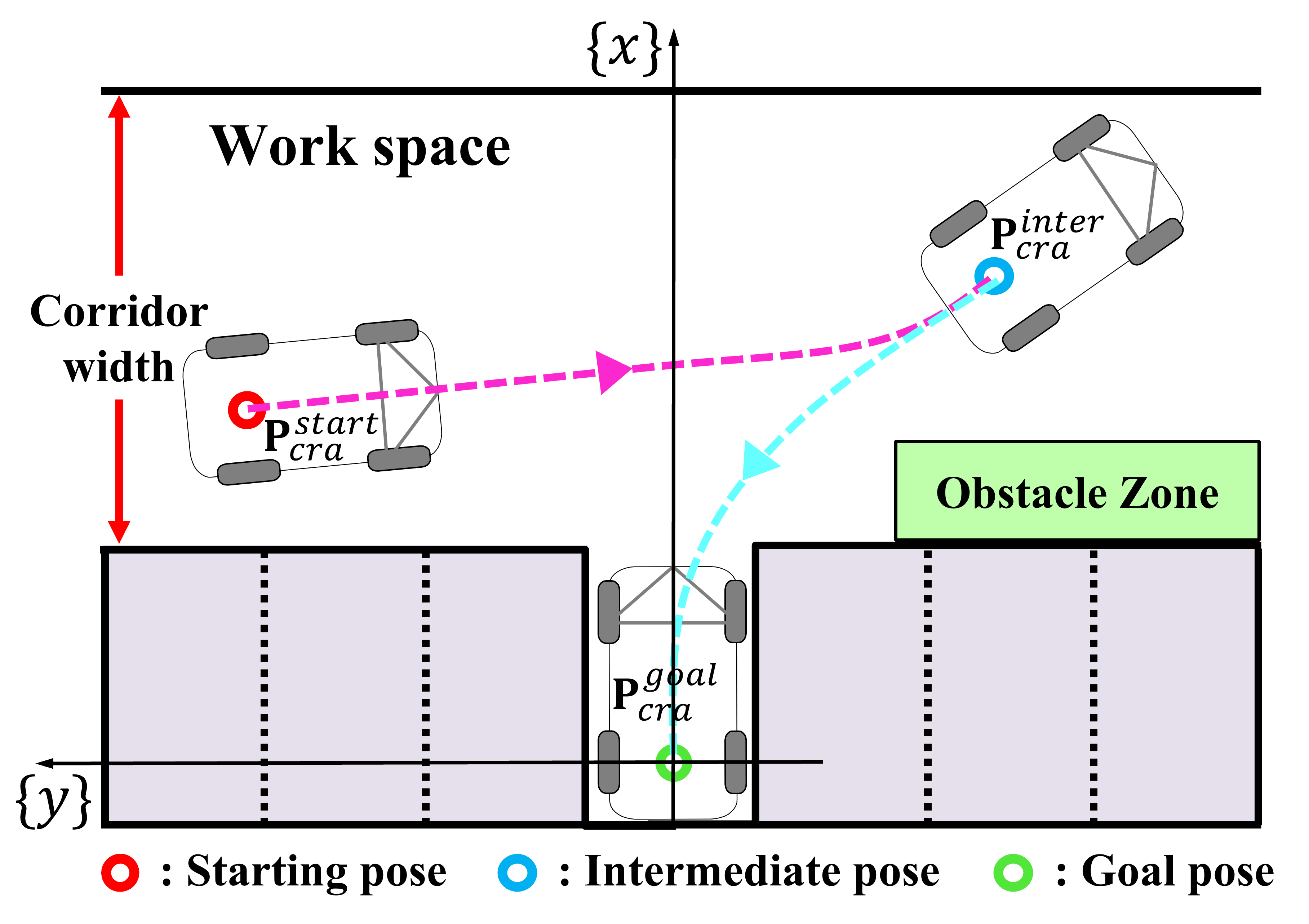}   
\caption{ There are three poses: starting pose ($\textbf{p}_{cra}^{start}$), intermediate pose ($\textbf{p}_{cra}^{inter}$) and goal pose ($\textbf{p}_{cra}^{goal}$).}
\label{fig:scenario}
\end{figure}

\begin{figure*}[t]
\centering
\includegraphics[width=1.0\textwidth]{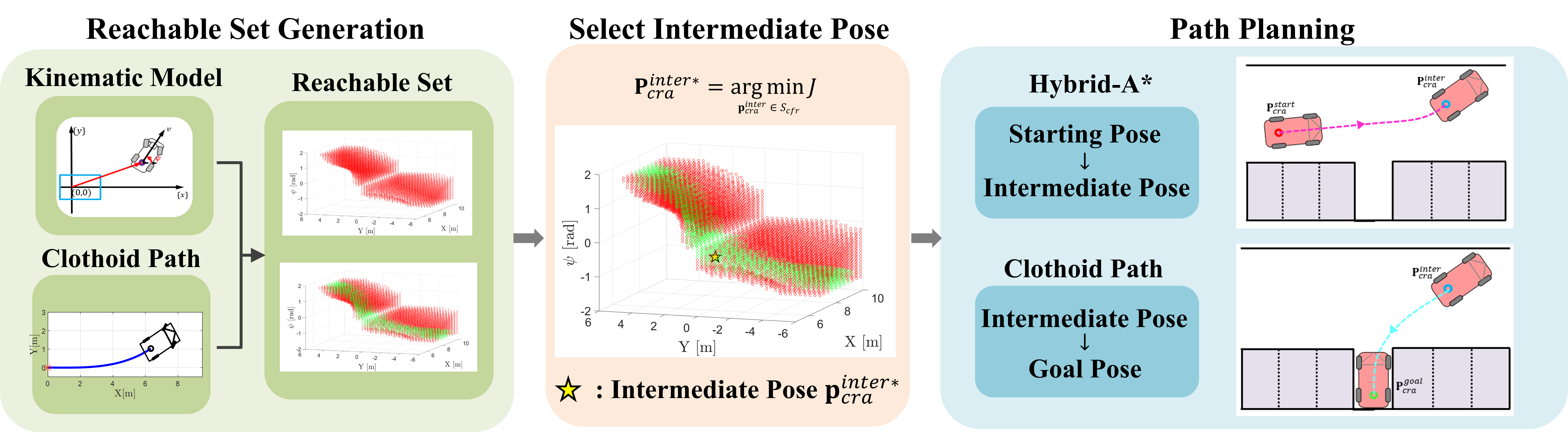}
\caption{The overall structure of the path planning. First, a collision-free reachable set is formed using the kinematic model and clothoid curve. Then, the cost function is utilized to select an appropriate intermediate pose. Subsequently, the path planning for the automated vertical parking system is conducted using the Hybrid-A* algorithm and clothoid-based method.}
\label{fig:overall structure_pathplanning}
\end{figure*}

\subsection{Problem Description}
The scenario of the automated parking system is illustrated in Fig.~\ref{fig:scenario}. We considered a parking lot with corridors of 7m and a minimum size of 6m. For each corridor width size, we evaluate the presence and absence of obstacles. Path planning from the starting pose $\textbf{p}_{cra}^{start}$ to the goal pose $\textbf{p}_{cra}^{goal}$ is required for the vehicle to move safely without collision to the parking spot. The first step for the path planning is to determine the appropriate intermediate pose $\textbf{p}_{cra}^{inter}$. Then, the path from $\textbf{p}_{cra}^{start}$ to $\textbf{p}_{cra}^{inter}$ can be generated. Finally, the reverse parking path from $\textbf{p}_{cra}^{inter}$ to $\textbf{p}_{cra}^{goal}$ is generated. Note that we focus exclusively on vertical parking scenarios.

\begin{assumption}
In this paper, we assume that we have prior knowledge of the workspace, which represents the area where the vehicle operates and carries out parking maneuvers. Furthermore, we have information about the pose of the ego vehicle relative to the final parking spot.
\end{assumption}

\section{Reachable Set-based Path Planning}
Implementing APS requires generating two paths: from the starting pose to the intermediate pose and from the intermediate pose to the goal pose. Determining the intermediate pose connecting the two paths is essential in integrating both path planning methods. To this end, we propose a reachable set of available intermediate poses. Figure~\ref{fig:overall structure_pathplanning} shows the overall structure of the proposed path planning. First, a reachable set is conducted using the kinematic model and the approximated clothoid path. Then, an appropriate intermediate pose is determined from the reachable set. Finally, we use Hybrid-A* for the first path that performs well in a narrow space, and the second path is obtained using a clothoid curve model, generating a smooth path.

\subsection{Approximated Clothoid-Based Parking Path Planning}
\label{subsec:clothoid}
This paper proposes a path generation approach that considers vehicle motion control towards the parking spot once the intermediate pose is determined. This approach aims to generate paths where all local paths are continuous and slowly varying curves. To this end, a clothoid can be used for smooth steering maneuvering. Given the arc length $s$ of a path, the curvature can be designed with the clothoid path construction rule as follows:
\begin{gather}
\kappa(s)=2c_2+6c_3s
\end{gather}
where $c_2$ and $c_3$ denote the path curvature at $s=0$ and its variation rate, respectively. However, implementing clothoid path planning in real-time can be computationally intensive. The arc length $s$ can be approximated to address this problem as the position $x$, assuming a small curvature. Then, using the integrated curvature, an approximated clothoid cubic polynomial tangent angle $\theta(\cdot)$ and curve model $f(\cdot)$~\cite{c16} can be derived as
\begin{gather}
\begin{split}
\label{eq:clothoid_curve}
&\theta(x) = c_1+2c_2x+3c_2x^2\\
&f(x) = c_0+c_1x+c_2x^2+c_3x^3
\end{split}
\end{gather}
where $c_0$ and $c_1$ represent the initial lateral offset and the initial orientation angle offset in the parking coordinates. This approximated model~\eqref{eq:clothoid_curve} can be utilized for the automated parking system with a virtual towing distance to prevent a numerical singularity problem, and the detailed process of this method is derived in~\cite{c10}.

In kinematic model it can be assumed that the vehicle velocity is small enough so that the velocity vector aligns with the direction of the wheel~\cite{c15}. Under steady-state conditions, assuming a constant low velocity, the vehicle has a circular motion with no sideslip for any of the tires, and a yaw rate is given by
\begin{gather}
\begin{split}
\dot{\psi}=\frac{v}{R_\kappa}=\frac{v}{l}\tan(\delta)
\end{split}
\end{gather}
where $R_{\kappa}$ is the turning radius determined by the curvature. Assuming that there are no model uncertainties and using $\kappa=1/R_{\kappa}$, the desired steering angle is obtained by

\begin{gather}
\begin{split}
\label{eq:desired steering}
\delta^d:=\tan^{-1}(\kappa l).
\end{split}
\end{gather}
%
%
%
Then, the desired state $\textbf{p}^{d}_{cra}=[x^d\;y^d\;\psi^d]^T$ can be obtained through \eqref{eq:discrete_kinematic_model} by taking the desired steering wheel angle $\delta^{d}$.

\subsection{Computation of Reachable Set}
\begin{table}[t]
\caption{Parameters of the vehicle and parking slot}
\label{tb:parameter}
\centering
\begin{tabular}{c|c|c}
\hline
\rowcolor[HTML]{FFFFFF}
\textbf{Parameter} & \textbf{Symbol} & \textbf{Value} \\ \hline
Corridor Length    & \textit{$l_{cl}$}     & 12{m}      \\
Corridor Width     & \textit{$l_{cw}$}     & 6{m}, 7{m} \\
Vehicle Length     & \textit{$l_{vl}$}     & 4.325{m}   \\
Vehicle Width      & \textit{$l_{vw}$}     & 1.890{m}   \\
Wheel Base         & \textit{l}      & 2.630{m}   \\
Rear Overhang      & \textit{$l_r$}   & 0.845{m}   \\
Parking Slot Length& \textit{$l_{sl}$}     & 5.5{m} \\
Parking Slot Width & \textit{$l_{sw}$}  & 2.9{m} \\ \hline
\end{tabular}
\end{table}
\begin{figure}[t]
\centering
\subfigure[][]{\includegraphics[width=0.4\textwidth]{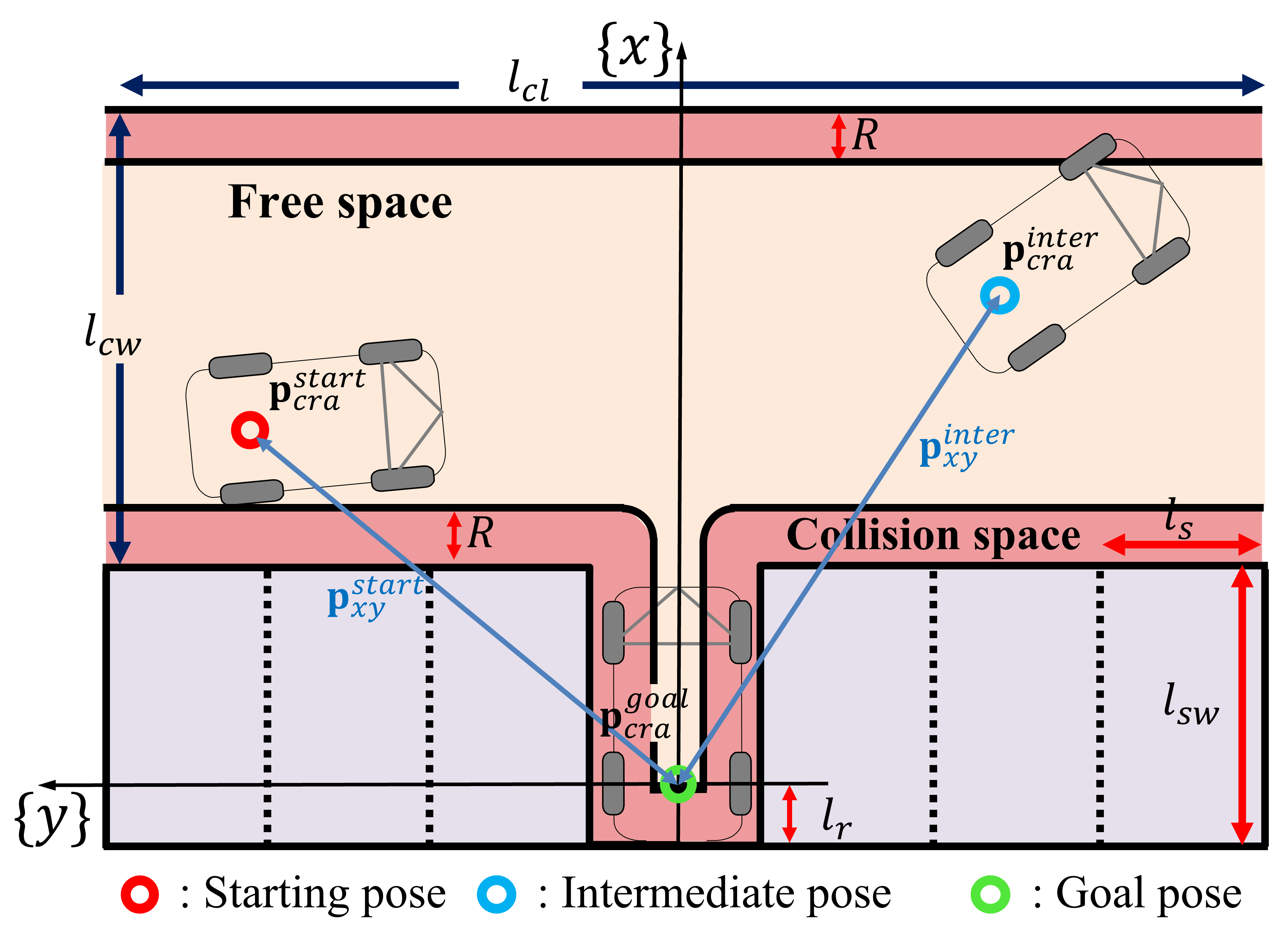}}
\subfigure[][]{\includegraphics[width=0.24\textwidth]{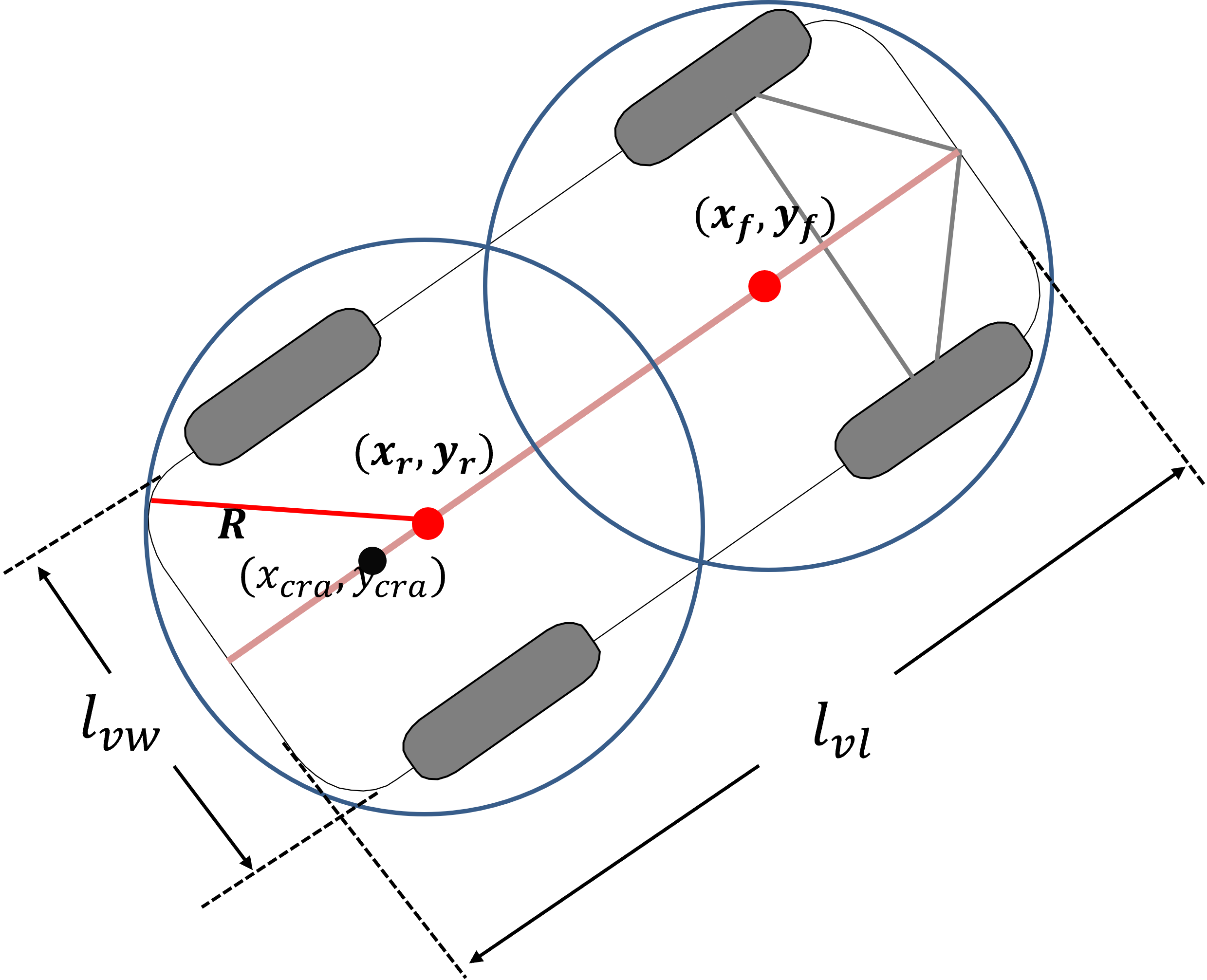}}   
\caption{(a) Visualize the free space and collision space within the parking lot. It also illustrates the length and width of the corridor and the size of the parking slot. (b) indicates parameters of the vehicle size. It also illustrates $(x_r,y_r)$, $(x_f, y_f)$, and radius $R$ that can define the free space and collision space.}
\label{fig:overall_scenario}
\end{figure}
\begin{algorithm}[t]
\caption{Collision-free reachable set $S_{cfr}$ generation}\label{reachable set algorithm}
\begin{algorithmic}[1]
\Require Grid set ($S_g$), Free space ($\mathcal{F}$)
\State $S_{r}=\emptyset$, $S_{cfr}=\emptyset$, $Collision=\emptyset$
\For{$i_{x} \gets 1$ to $N_{x}$}
\For{$i_{y} \gets 1$ to $N_{y}$}
\For{$i_{\psi} \gets 1$ to $N_{\psi}$}
\State $\textbf{p}_{cra}^{d}=[x_{i_{x}},\,y_{i_{y}},\,\psi_{i_{\psi}}]^T$
\While{$\textbf{p}_{cra}^{d}(1)>0$}
\State Obtain $\delta^{d}$ using Eq.~\eqref{eq:desired steering}
\State Obtain $[x^{d},\,y^{d},\,\psi^{d}]^T$ using Eq.~\eqref{eq:discrete_kinematic_model}
\State Update $\textbf{p}_{cra}^{d}=[x^{d},\,y^{d},\,\psi^{d}]^T$
\If{$(x_{r},y_{r}),(x_{f},y_{f})\in \mathcal{F}$}
\State $Collision=\emptyset$
\Else
\State Add 1 to $Collision$
\EndIf
\EndWhile
\If{$|\textbf{p}_{cra}^{d}(2)|\leq \epsilon_{y}$ and $|\textbf{p}_{cra}^{d}(3)|\leq \epsilon_{\psi}$}
\State Add $[x_{i_{x}}\,y_{i_{y}}\,\psi_{i_{\psi}}]^{T}$ to $S_{r}$
\If{$Collision=\emptyset$}
\State Add $[x_{i_{x}}\,y_{i_{y}}\,\psi_{i_{\psi}}]^{T}$ to $S_{cfr}$
\EndIf
\EndIf
\EndFor
\EndFor
\EndFor
\end{algorithmic}
\label{alg:reachable_alg}
\end{algorithm}
Selecting an appropriate intermediate pose is crucial to reach the goal pose effectively. To this end, we define a reachable set and a collision-free reachable set.
\begin{definition}
  Reachable set, $S_r$, is a 3-dimensional set of intermediate poses, $\textbf{p}_{cra}^{inter}$, where the vehicle can generate a vertical parking path to the goal pose with a single reverse maneuver.
\end{definition}
\begin{definition}
  Collision-free reachable set, $S_{cfr}$, is a subset generated by considering collision avoidance in the reachable set, $S_r$.
\end{definition}

The first step to compute the reachable set is to define the grid set, $S_g$, by dividing the parking corridor into grids with specific interval. The grid range can be defined using upper limits and lower limits of the vehicle states as $\uubar{{x}}=(l_{sl}-l_r)+l_{vl}/2$, $\bar{x} = (l_{sl}-l_r)+l_{cw}-l_{vl}/2$, $\uubar{{y}}=-l_{cl}/2$, $\bar{y} = l_{cl}/2$, $\uubar{{\psi}}=-\pi/2$, and $\bar{\psi} = \pi/2$ where all parameters are described in Table~\ref{tb:parameter}. Then,
a set $S_g$, which has finite 3-dimensional grid points, is defined as follows:
\begin{gather}
\begin{split}
S_g := \{[x_{i_{x}},y_{i_{y}},\psi_{i_{\psi}}]^{T}|&\uubar{x}\leq x_{i_{x}}\leq\bar{x}\,\,\, \textrm{for}\,\,\,i_{x}=1,\hdots,N_{x},\\ &\uubar{y}\leq y_{i_{y}}\leq\bar{y} \,\,\, \textrm{for}\,\,\, i_{y}=1,\hdots,N_{y},\\ &\uubar{\psi}\leq \psi_{i_{\psi}}\leq\bar{\psi}\,\,\,\textrm{for}\,\,\, i_{\psi}=1,\hdots,N_{\psi}\}
\end{split}
\end{gather}
where the number of grid points is $N_{x}\times N_{y}\times N_{\psi}$.
After the generation $S_{g}$, the clothoid-based path is generated by computing $\textbf{p}_{cra}^{d}\in S_{g}$ until the parking process is completed. If the $y$ and $\psi$ values of the final pose are smaller than the pre-defined threshold values $\epsilon_y, \epsilon_{\psi}$, the $\textbf{p}_{cra}^{d}$ of the first step, which corresponds to the intermediate pose, is added in the reachable set, $S_r$. However, this process does not consider collision checking based on the surrounding environment of the parking lot.
The reachable set should have some constraints to consider the collision. To compute the collision-free reachable set, two points $(x_r,y_r)$ and $(x_f,y_f)$ considering the size of the vehicle are defined as follows:
\begin{gather}
\label{eq:xr,xf,yr,yf}
\begin{split}
&x_r = x^d + (\frac{l_{vl}}{4}-l_r)\cos(\psi^d)\\
&y_r = y^d + (\frac{l_{vl}}{4}-l_r)\sin(\psi^d)\\
&x_f = x^d + (\frac{3l_{vl}}{4}-l_r)\cos(\psi^d)\\
&y_f = y^d + (\frac{3l_{vl}}{4}-l_r)\sin(\psi^d).
\end{split}
\end{gather}
Then, the radius $R$ of the circle around \eqref{eq:xr,xf,yr,yf} is defined by
\begin{gather}
\label{eq:radius}
\begin{split}
R = \sqrt{(\frac{l_{vl}}{4})^2+(\frac{l_{vw}}{2})^2}
\end{split}
\end{gather}
to deal with the collision as shown in Fig.~\ref{fig:overall_scenario}. Here, two points $(x_r,y_r)$ and $(x_f,y_f)$ are the coordinates of the quarter point and three-quarter point of the vehicle center line, based on the center of the rear axle. Thus, it is possible that we can cover all vehicle body with two circles using \eqref{eq:xr,xf,yr,yf} and \eqref{eq:radius} as shown in Fig.~\ref{fig:overall_scenario} (b).

Let a free space $\mathcal{F}$ refers to obstacle-free areas where a vehicle can safely navigate without any collisions~\cite{c5}. By using ~\eqref{eq:xr,xf,yr,yf} and~\eqref{eq:radius}, the free space can be obtained by adding a margin of $R$ around restricted areas and obstacles within the workspace. As a result, in each step of constructing the reachable set, if the coordinates $(x_r,y_r)$ and $(x_f,y_f)$ obtained from $\textbf{p}_{cra}^{d}$ are within the free space, the path is collision-free. Then, the $\textbf{p}_{cra}^{d}$ of the first step, which corresponds to the intermediate pose, is added in the collision-free reachable set, $S_{cfr}$. The overall process generating the reachable set is presented in \textbf{Algorithm \ref{alg:reachable_alg}}. If the information about the environment and surrounding static obstacles in the parking lot is known, one can calculate and store the collision-free reachable set offline, tailored to that specific parking lot. With this, it becomes possible to achieve safe parking.
\begin{figure}[t]
\centering
\subfigure[][]{\includegraphics[width=0.38\textwidth]{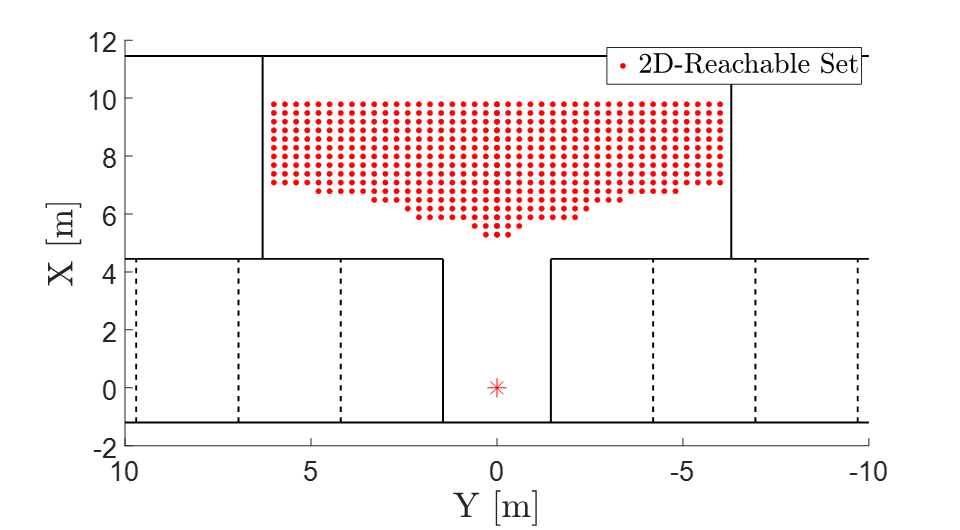}}
\subfigure[][]{\includegraphics[width=0.38\textwidth]{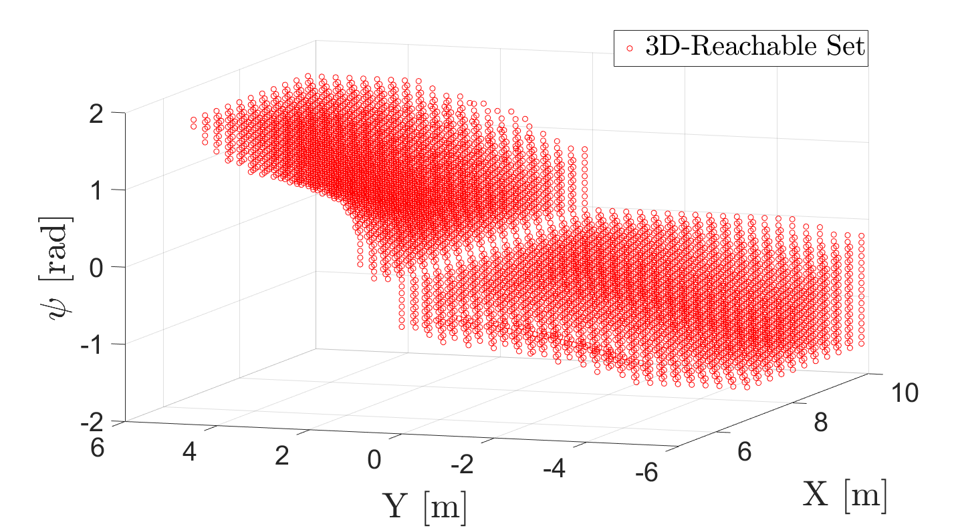}}
\caption{(a) 2-dimensional reachable set, (b) 3-dimensional reachable set. The red points represent intermediate poses that allow for generating a single reverse path to the final parking spot.}
\label{fig:reachableset_no_constraint}
\end{figure}

\begin{figure}[t]
\centering
\subfigure[][]{\includegraphics[width=0.38\textwidth]{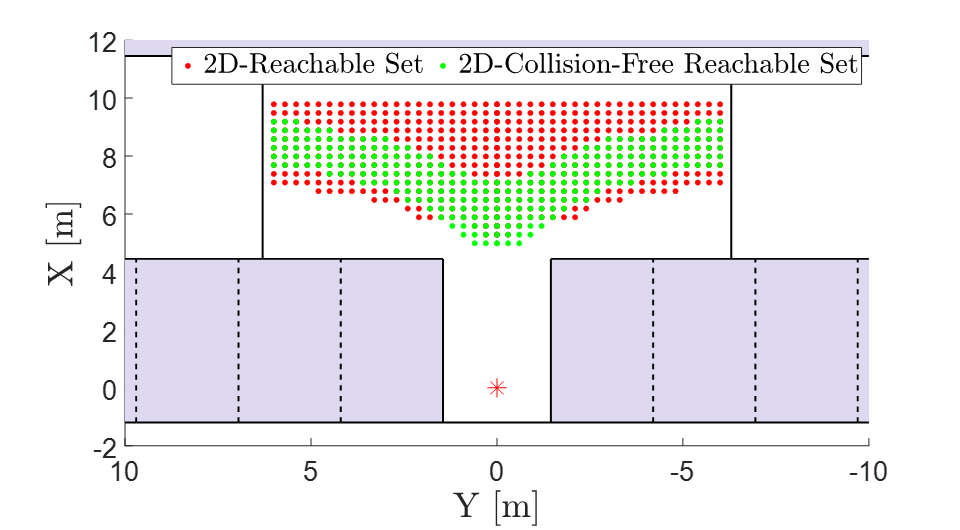}}
\subfigure[][]{\includegraphics[width=0.38\textwidth]{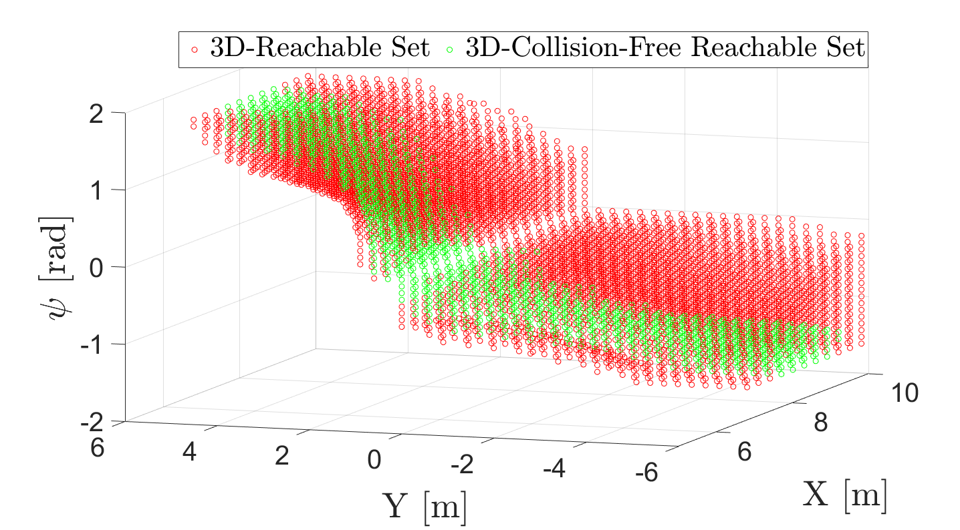}}
\caption{(a) 2-dimensional collision-free reachable set (b) 3-dimensional collision-free reachable set. Purple area in (a) means restricted area. Due to the restricted area, the range of green points has been reduced.}
\label{fig:reachableset_with_constraint}
\end{figure}

The reachable set is shown in Fig.~\ref{fig:reachableset_no_constraint}. The intermediate poses included in $S_r$ visualized in two-dimensional are represented in Fig.~\ref{fig:reachableset_no_constraint} (a). The red points represent the intermediate poses that can be reached at each position using the clothoid-based path generation method, leading to the goal pose. In Fig.~\ref{fig:reachableset_no_constraint} (b), which is presented in a three-dimensional, it can be observed that the number of possible orientations varies depending on the position. If there are restricted areas (purple area) in the parking lot, such as Fig.~\ref{fig:reachableset_with_constraint} (a), the intermediate poses included in $S_{cfr}$ that can generate collision-free paths are visualized as green points. When considering collision avoidance, it is evident that the range of green points significantly reduces. Also, in Fig.~\ref{fig:reachableset_with_constraint} (b), we can observe that the green points vary depending on the orientation for the same position. The changes in the shape of the reachable set in the presence of obstacles will be illustrated in Section~\ref{sec:simulation_results}.

By selecting a single grid point among the numerous grid points of the collision-free reachable set, it can be designated as an intermediate pose $\textbf{p}_{cra}^{inter}$. Finally, if the vehicle reaches the intermediate pose $\textbf{p}_{cra}^{inter}$, it becomes possible to achieve collision-free vertical parking with a smooth steering wheel angle using clothoid-based path planning.
\subsection{Select Intermediate Pose in Reachable Set}
In this subsection, we will explain how to select the appropriate intermediate pose $\textbf{p}_{cra}^{inter*}$ among numerous grid points in the collision-free reachable set $S_{cfr}$. The intermediate pose can be chosen based on the relationship between the starting pose $\textbf{p}_{cra}^{start}$ and $\textbf{p}_{cra}^{inter}$, as well as the relationship between $\textbf{p}_{cra}^{inter}$ and the goal pose $\textbf{p}_{cra}^{goal}$. To this end, we introduce the cost function with weight parameters $\alpha_1,\hdots,\alpha_4$ defined as
\begin{gather}
\begin{split}
J=\alpha_1J_1 + \alpha_2J_2 + \alpha_3J_3 + \alpha_4J_4
\end{split}
\end{gather}
where
\begin{gather*}
\begin{split}
&J_1 = |\textbf{p}_{cra}^{start}(3)-\textbf{p}_{cra}^{inter}(3)|\\
&J_2 = \|\textbf{p}_{xy}^{start}-\textbf{p}_{xy}^{inter}\|_{2}\\
&J_3 = \|\textbf{p}_{xy}^{goal}-\textbf{p}_{xy}^{inter}\|_{2}\\
&J_4 = |\psi^{pref}-\textbf{p}_{cra}^{inter}(3)|.
\end{split}
\end{gather*}

\begin{figure*}[t]
\centering
\subfigure[][7m case without obstacle zone]{\includegraphics[width=0.305\textwidth]{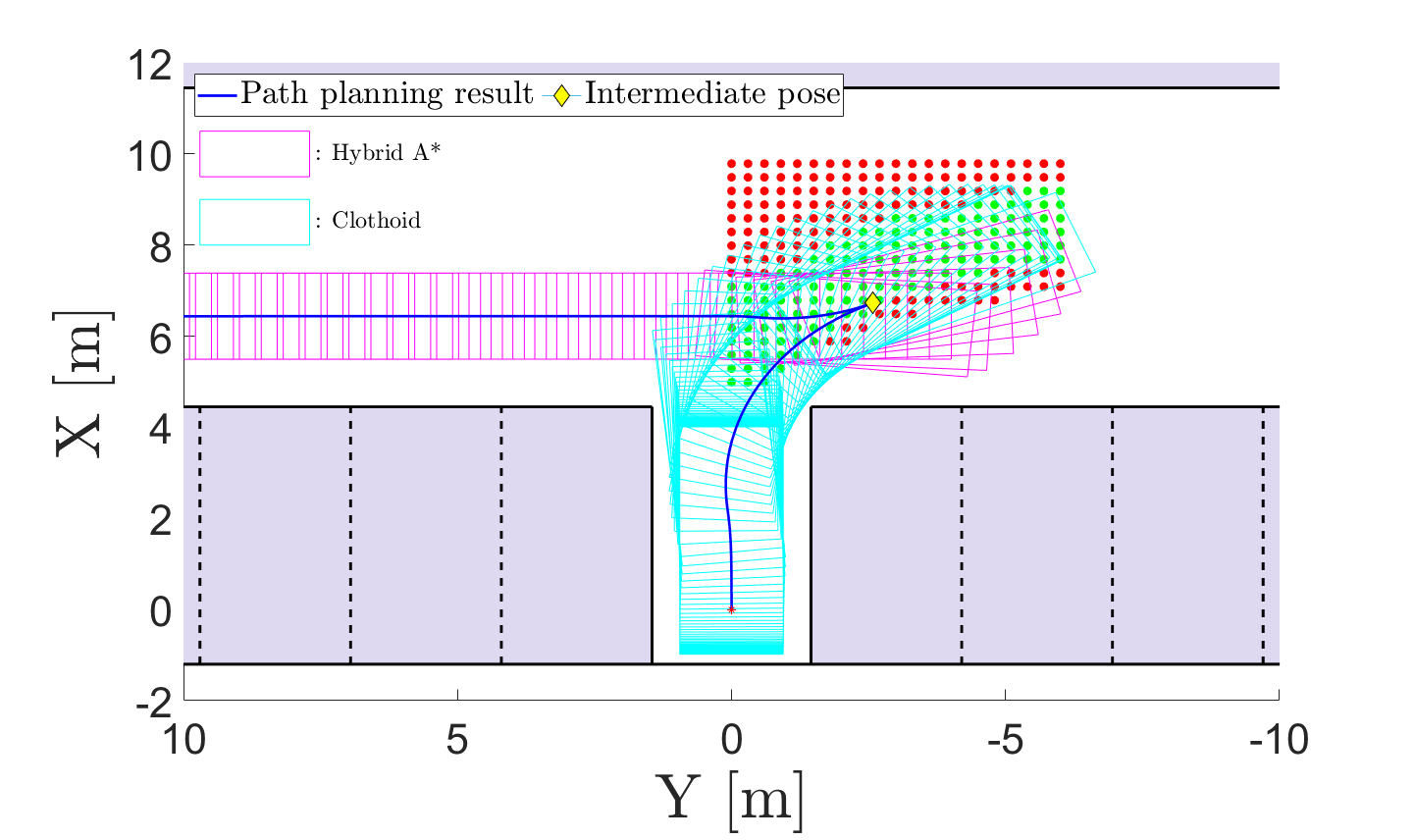}}
\subfigure[][7m case with obstacle zone (top)]{\includegraphics[width=0.305\textwidth]{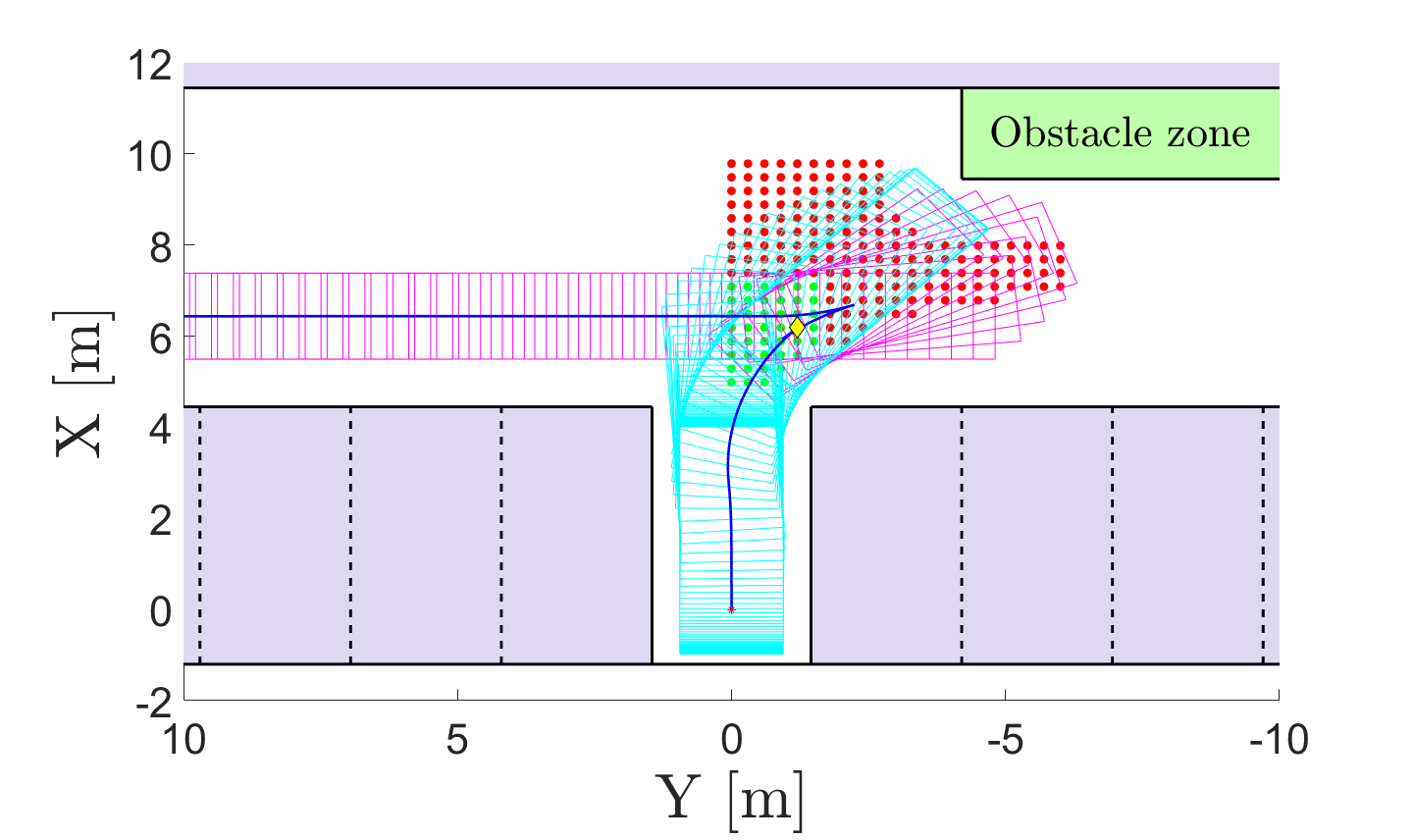}}
\subfigure[][7m case with obstacle zone (bottom)]{\includegraphics[width=0.305\textwidth]{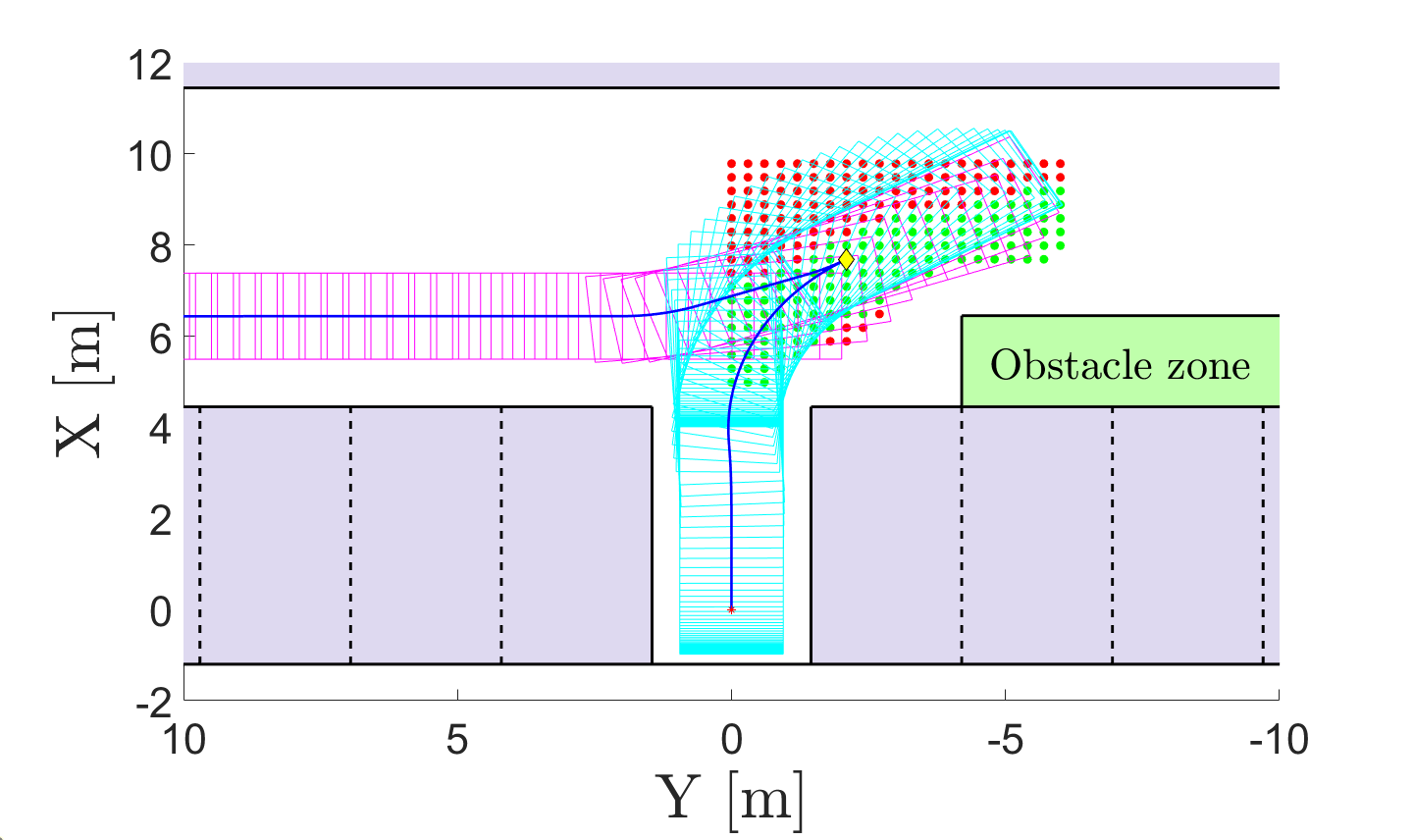}}
\subfigure[][6m case without obstacle zone]{\includegraphics[width=0.305\textwidth]{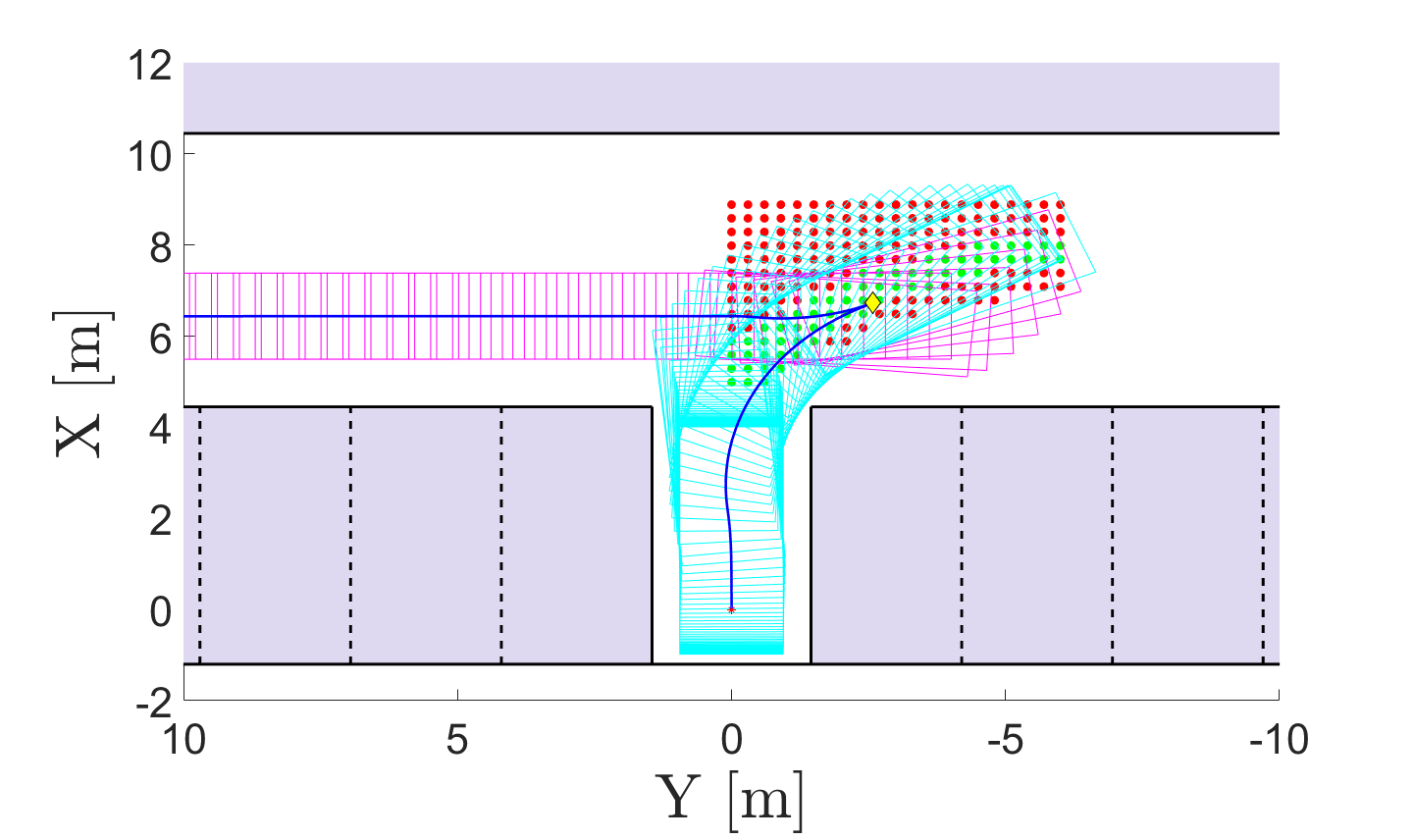}}
\subfigure[][6m case with obstacle zone (top)]{\includegraphics[width=0.305\textwidth]{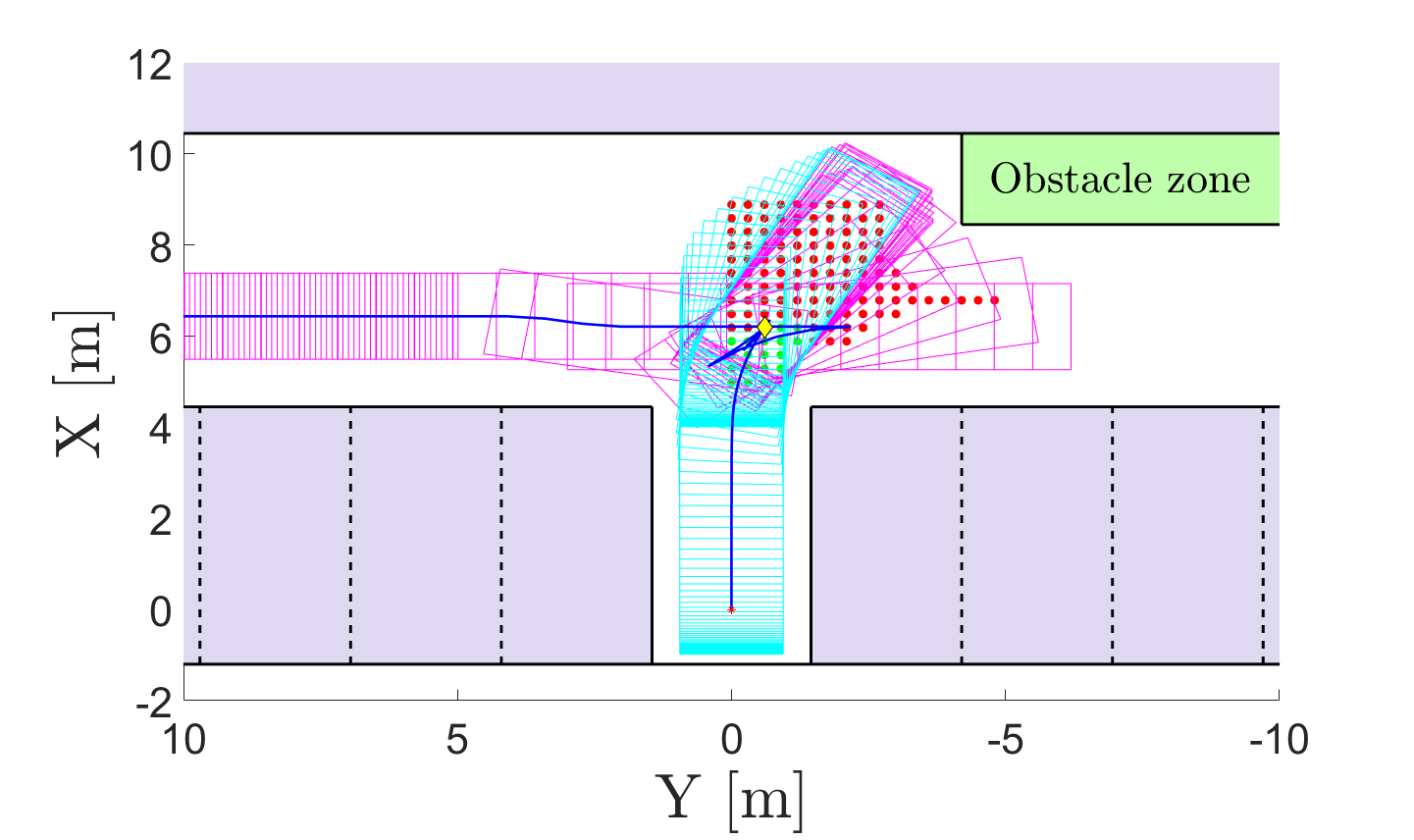}}
\subfigure[][6m case with obstacle zone (bottom)]{\includegraphics[width=0.305\textwidth]{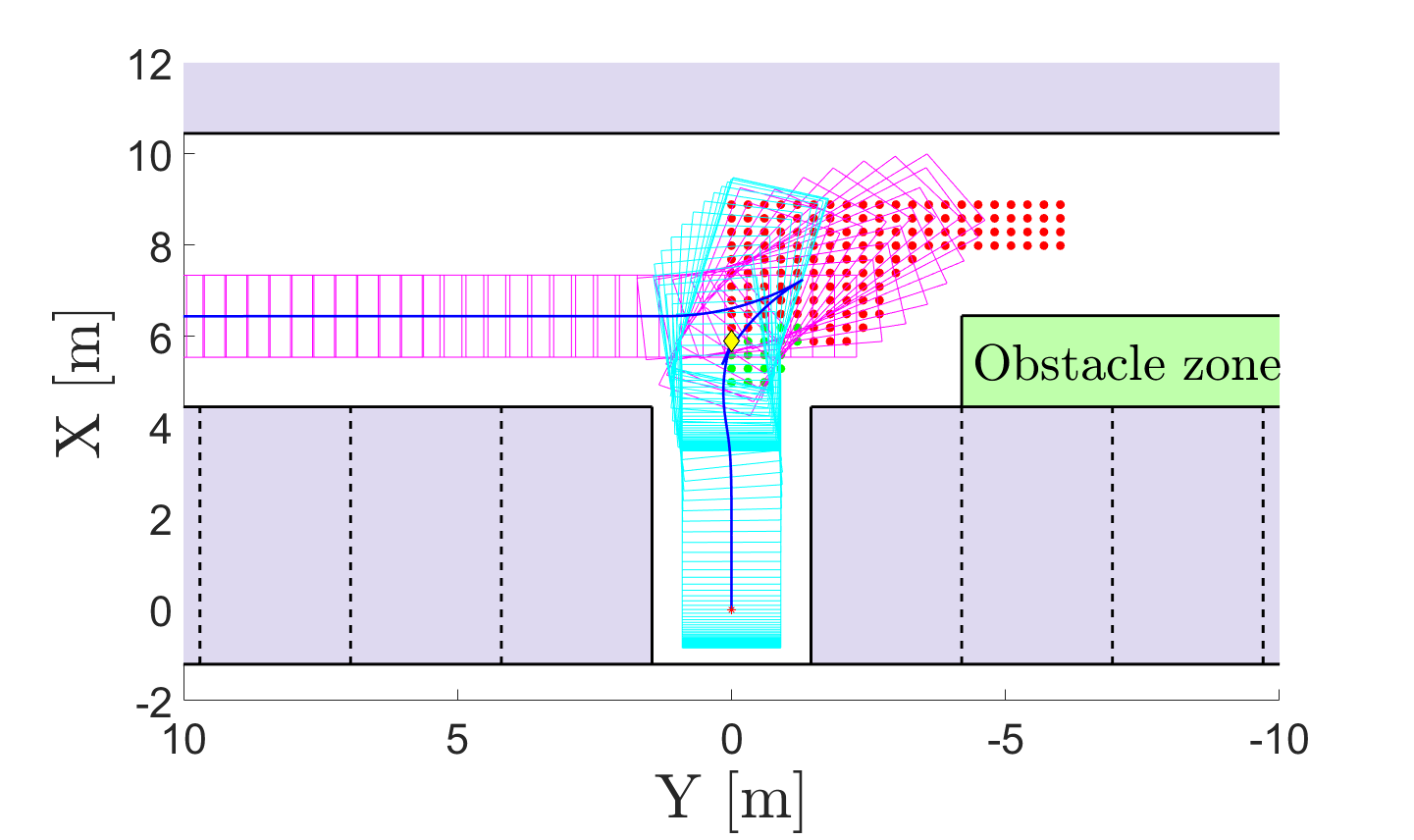}}
\caption{Path planning results with various scenarios. (a)-(c) are cases with the width of 7m corridor and (d)-(e) are cases with the width of 6m corridor. ((a),(c),(d) occurred once for gear shift, (b),(f) occurred three times, and (e) occurred five times.)}
\label{fig:pathplanning_result}
\end{figure*}
Here, $J_1$ is a cost that ensures minimal steering changes from the starting pose to the intermediate pose. $J_2$ and $J_3$ are costs that prevent unnecessary driving distances from the intermediate pose to the starting pose and goal pose, respectively. $J_4$ is a cost representing the difference between the vehicle orientation in the intermediate pose and the driver's preference with the pre-defined orientation $\psi^{pref}$. To achieve safe parking, the orientation of the intermediate pose is considered the most crucial among all costs. Therefore, we assigned more weight to $\alpha_4$ and performed tuning to match the units of each cost. Then, to select the appropriate intermediate pose dependent on $\psi^{pref}$ in the reachable set, the minimization problem is given by
\begin{gather}
\begin{split}
\textbf{p}_{cra}^{inter*}=\argmin_{\textbf{p}_{cra}^{inter}\in S_{rc}}{J}(\psi^{pref}).
\end{split}
\end{gather}
After determining $\textbf{p}_{cra}^{inter*}$, we can generate the reference path for the vehicle from the $\textbf{p}_{cra}^{start}$ to $\textbf{p}_{cra}^{inter*}$ using a general Hybrid-A* algorithm. Furthermore, we already described in Section~\ref{subsec:clothoid} that $\textbf{p}_{cra}^{inter*}$ can form the collision-free vertical parking path to the $\textbf{p}_{cra}^{goal}$. Finally, the path from the starting pose to the goal pose has been generated.

\section{Simulation Results}
\label{sec:simulation_results}
\begin{figure}[b]
\centering
\includegraphics[width=0.41\textwidth]{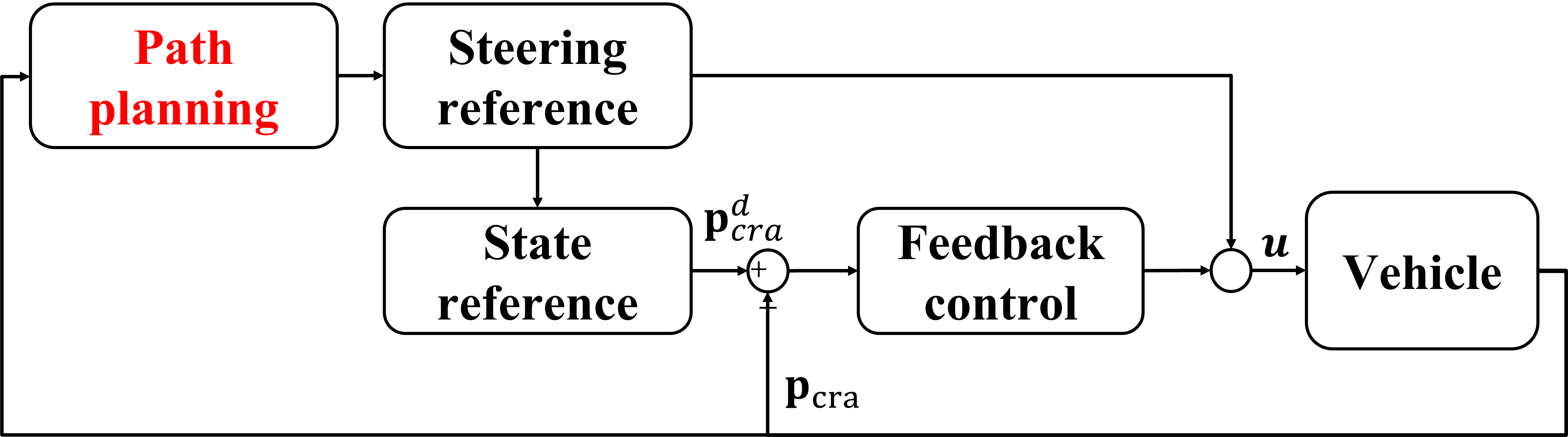}
\caption{The overall structure of the tracking process. We use a feedforward controller and feedback controller to track the reference path.}
\label{fig:overall_structure_tracking}
\end{figure}

\subsection{Path Planning Result}

We conducted path planning simulations for six scenarios to verify the proposed method in MATLAB. The parameters of the vehicle used in the simulation are shown in Table~\ref{tb:parameter}. First, a simulation was performed to evaluate the proposed approach using a parking lot with a corridor size of 7m, which is typical for parking lots. Furthermore, the areas within the parking lot where the probability of static obstacles is high are designated as obstacle zones. Fig.~\ref{fig:pathplanning_result} (a), (b), and (c) shows the varying reachable set, influenced by the location and presence of the obstacle zone within a 7m corridor. Generated path results demonstrate successful path planning that avoids collisions with obstacles during parking. The results obtained for the corridor size of 6m are represented in Fig.~\ref{fig:pathplanning_result} (d), (e), and (f). Compared to the 7m scenario, we can observe that the area of the collision-free reachable set, represented by the green dots, shrinks. This shrinking of the green dots area can be interpreted as a result of the limited space in the parking lot environment, leading to reduced free space. Furthermore, due to the limited space, multiple gear shifts are required during the parking maneuver but still generate a collision-free path.

\subsection{Tracking Result}

\begin{figure}[t]
\centering
\subfigure[][Tracking result]{\includegraphics[width=0.42\textwidth]{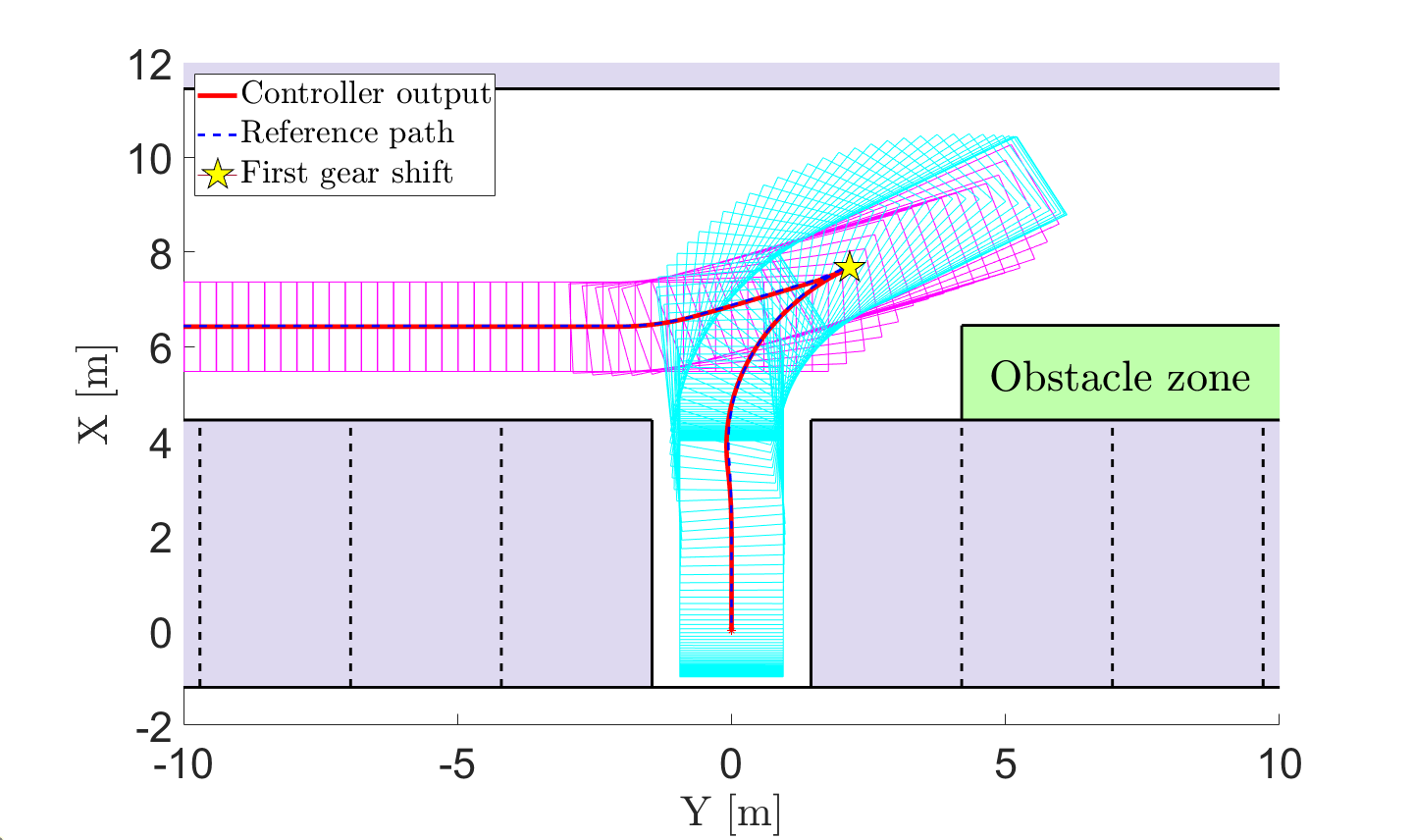}}
\subfigure[][Steering wheel angle]{\includegraphics[width=0.42\textwidth]{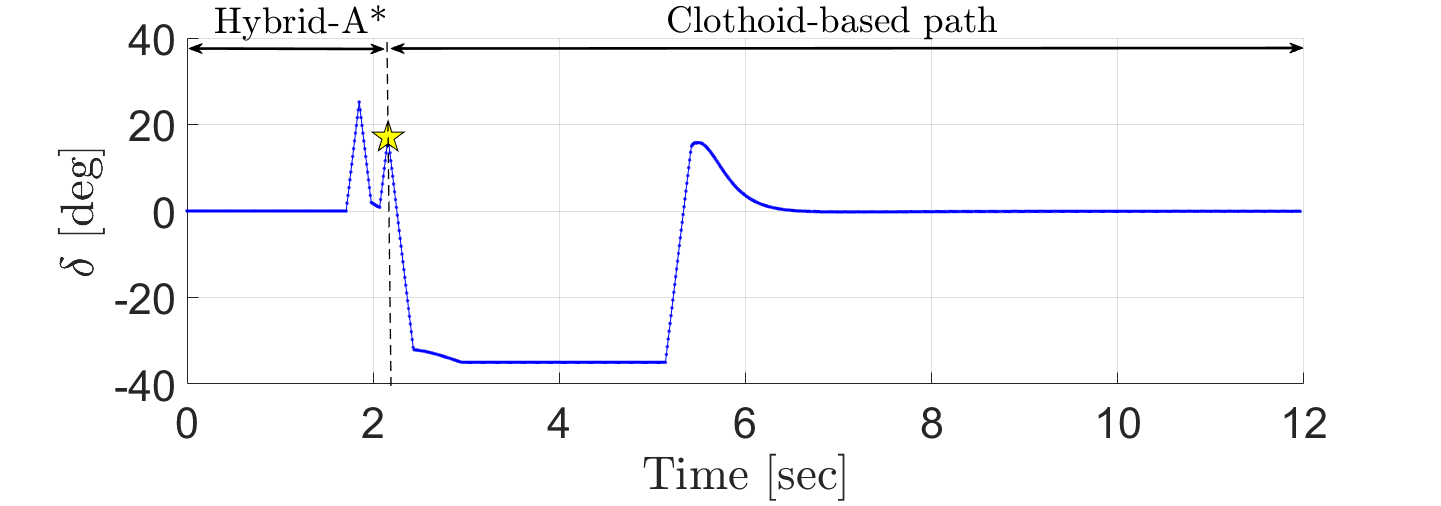}}
\caption{Tracking performance result for 7m case. (a) tracking result with reference path, (b) steering wheel angle result. The path is divided into a Hybrid-A* path and a clothoid-based path, both before and after reaching the intermediate pose.}
\label{fig:control_7m}
\end{figure}

\begin{figure}[t]
\centering
\subfigure[][Tracking result]{\includegraphics[width=0.42\textwidth]{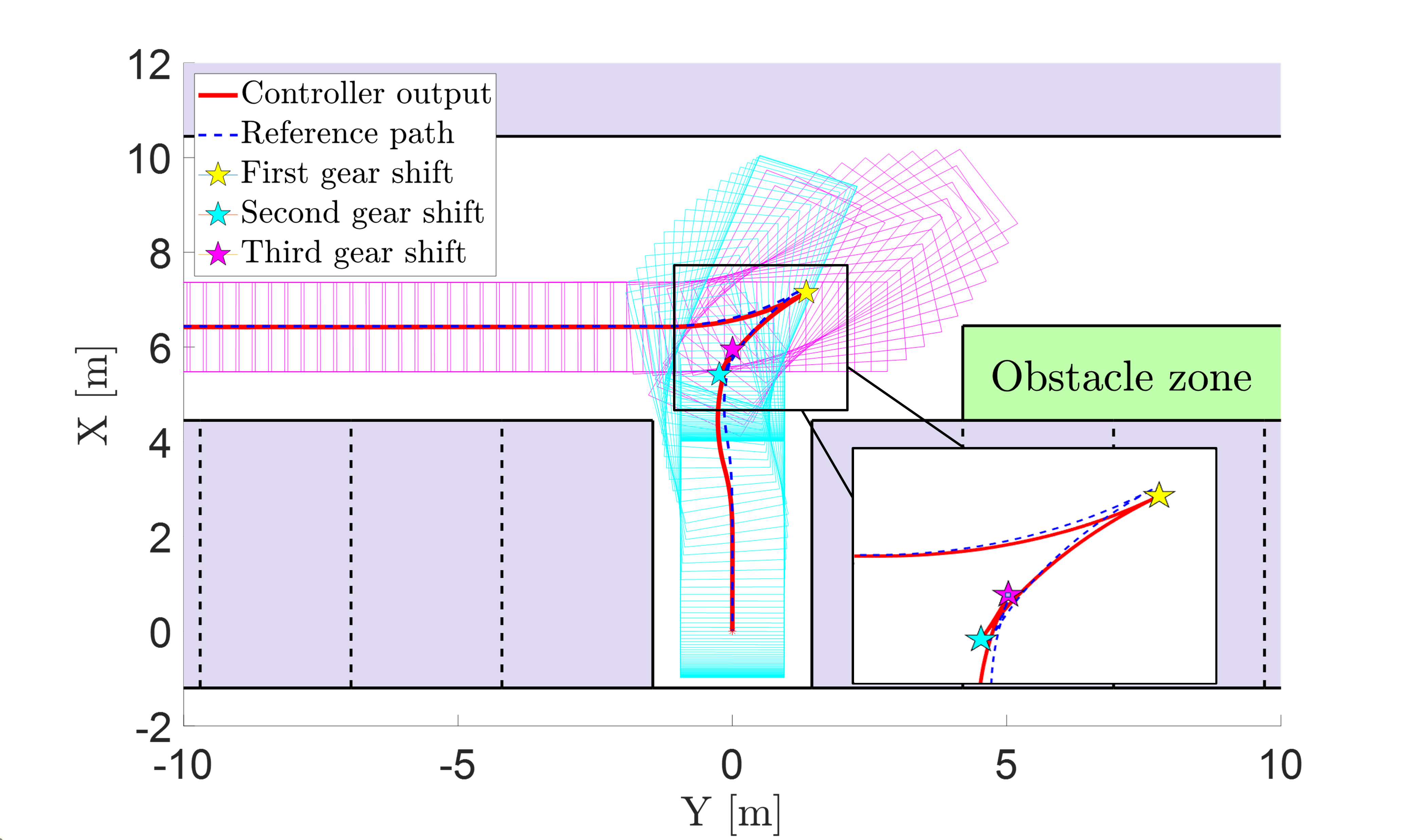}}
\subfigure[][Steering wheel angle]{\includegraphics[width=0.42\textwidth]{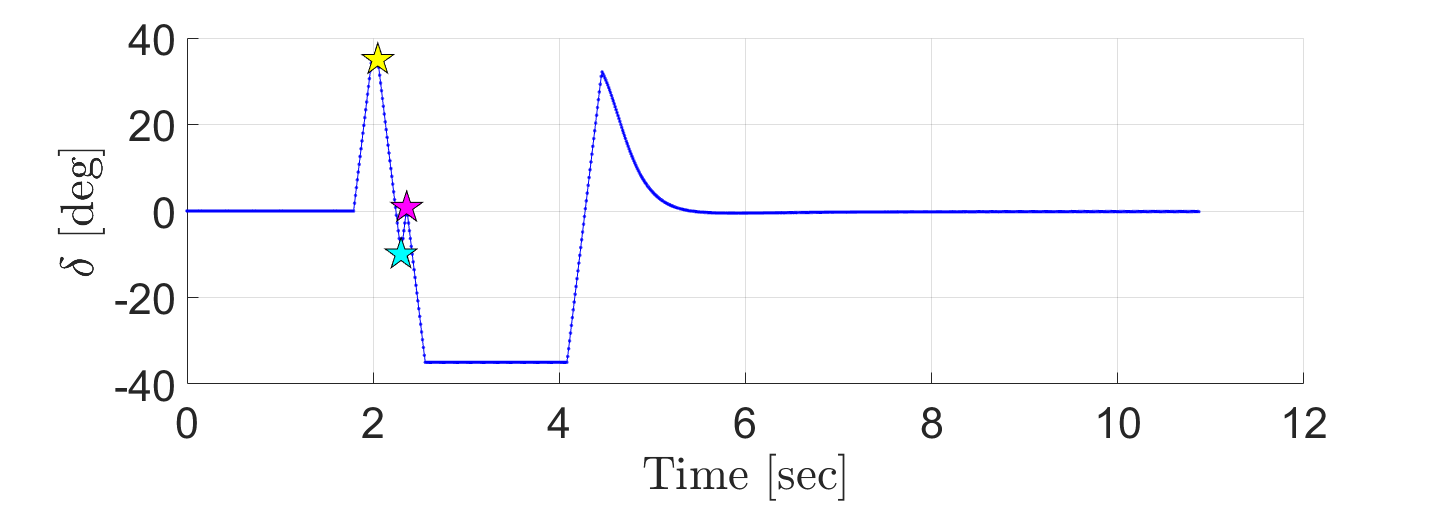}}
\caption{Tracking performance result for 6m case. (a) tracking result with reference path, (b) steering wheel angle result. The path is divided into a Hybrid-A* path and a clothoid-based path, both before and after reaching the intermediate pose.}
\label{fig:control_6m}
\end{figure}

In order to consider the uncertainty of the vehicle model due to the external environment, we added bounded disturbances into the plant model and conducted simulations. The overall structure of the tracking process is illustrated in Fig.~\ref{fig:overall_structure_tracking}. We designed a feedforward controller to track the generated path and used a feedback controller to compensate for disturbances. During the simulation, we applied a restriction on the steering rate of the vehicle. The control simulation results for the case where the corridor size is 7m, and the obstacle zone is located at the bottom are represented in Fig.~\ref{fig:control_7m}. We can observe that the vehicle closely follows the reference path and reaches the final parking spot with a single gear shift.
Additionally, examining the steering wheel angle after the gear shift occurs, the generated path is based on clothoid so that smooth steering identifies. The results for the case where the obstacle remains in the same position but the corridor width is reduced to 6m are shown in Fig.~\ref{fig:control_6m}. It can be observed that there are three gear shifts due to the reduced space of the free space. Also, because we did not consider that typical vehicles align their wheels during gear shifting, deviations occurred between the vehicle's actual and reference paths. However, the vehicle successfully enters the parking spot without collision, and the steering wheel angle results also show smooth steering after the final gear shift. To validate the path tracking performance, we calculated the RMSE and maximum error of the tracking error between the reference path $\textbf{p}_{cra}^{d}$ and state $\textbf{p}_{cra}$. In the 7m case, the lateral position and orientation RMSE were 0.02m and 0.006rad, maximum error was 0.26m and 0.05rad, respectively. While in the 6m case, RMSE was 0.06m and 0.01rad, and maximum error was 0.3m and 0.17rad, respectively. A video clip can be seen at https://youtu.be/c3PeLBYq3uk in which Fig.~\ref{fig:pathplanning_result} (a), (b), (f) and Fig.~\ref{fig:control_6m}, respectively.

\section{CONCLUSIONS}
This paper proposed a reachable set for an automated parking system. We construct the reachable set using a vehicle kinematic model and a clothoid-based path planning method. The reachable set consists of intermediate poses where the vehicle can perform vertical parking with a single reverse maneuver. To obtain a collision-free reachable set, we defined free space based on the shape of the parking lot and the surrounding environments. Then, within the reachable set containing numerous points, we selected an intermediate pose by the cost function. Finally, generate the automated parking path using Hybrid-A* algorithm and clothoid-based methods. We conducted path planning simulations for six scenarios and confirmed that safe and collision-free paths were generated. By applying disturbances to the plant model and controlling it, we showed that the parking was completed without collision through smooth steering from the intermediate pose to the goal pose.

\addtolength{\textheight}{-12cm}   




\end{document}